\documentclass{article}

\PassOptionsToPackage{numbers, compress}{natbib}

\usepackage[main, final]{neurips_2026}

 \usepackage{amsmath}
\usepackage[utf8]{inputenc} 
\usepackage[T1]{fontenc}    
\usepackage{hyperref}       
\usepackage{url}            
\usepackage{booktabs}       
\usepackage{amsfonts}       
\usepackage{amssymb}        
\usepackage{nicefrac}       
\usepackage{microtype}      
\usepackage{xcolor}         
\usepackage{graphicx}
\usepackage{tabularx}       
\usepackage{bbm}            
\usepackage{multirow}
\usepackage{makecell}
     \usepackage{pdfpages}
     \usepackage{pgffor}
     \usepackage{hyperref}
\usepackage{mdframed}
\usepackage{listings}
\usepackage{xcolor}
\usepackage{booktabs}
\usepackage{makecell}
\newmdenv[
  linecolor=gray!50, backgroundcolor=gray!5,
  innertopmargin=4pt, innerbottommargin=4pt,
  innerleftmargin=8pt, innerrightmargin=8pt,
  skipabove=6pt, skipbelow=6pt
]{promptbox}
\lstdefinestyle{promptstyle}{
  basicstyle=\ttfamily\footnotesize,
  breaklines=true, breakatwhitespace=true,
  columns=fullflexible, keepspaces=true,
  frame=single, framesep=4pt, rulecolor=\color{gray!50},
  backgroundcolor=\color{gray!5},
  showstringspaces=false, aboveskip=4pt, belowskip=4pt
}
\usepackage{multirow}
\usepackage{makecell}
\usepackage{booktabs}
\lstdefinestyle{prompt}{
  basicstyle=\ttfamily\footnotesize,
  breaklines=true, breakatwhitespace=true,
  columns=fullflexible, keepspaces=true,
  frame=single, framesep=4pt, rulecolor=\color{gray!50},
  backgroundcolor=\color{gray!5},
  showstringspaces=false, aboveskip=4pt, belowskip=4pt
}

\definecolor{cfg-bg}{HTML}{FAFAFA}
\definecolor{cfg-nt}{HTML}{7C3AED}       
\definecolor{cfg-t}{HTML}{0E7490}        
\definecolor{cfg-op}{HTML}{B91C1C}       
\definecolor{cfg-comment}{HTML}{9CA3AF}  
\definecolor{cfg-str}{HTML}{047857}      
\definecolor{cfg-frame}{HTML}{E5E7EB}
\usepackage{titlesec}
\titlespacing*{\subsubsection}{0pt}{6pt}{3pt}
\titlespacing*{\subsection}{0pt}{6pt}{4pt}
 \titlespacing*{\section}{0pt}{6pt}{8pt}

\lstdefinestyle{cfgstyle}{
  backgroundcolor=\color{cfg-bg},
  basicstyle=\ttfamily\small\linespread{1.08}\selectfont,
  keywordstyle=\color{cfg-nt}\bfseries,
  keywordstyle=[2]\color{cfg-t}\bfseries,
  commentstyle=\color{cfg-comment}\itshape,
  stringstyle=\color{cfg-str},
  morekeywords={Plugin,MethodSpec,Pipeline,DataMgr,DataParser,
    Model,Optimizers,BaseOpt,CustomOpt,Viewer,Package,
    EntryPoints,Modification,ExtPoint},
  morekeywords=[2]{TrainerConfig,VanillaPipelineConfig,
    FullImageDMConfig,DepthDMConfig,FeatureDMConfig,CustomDMConfig,
    NerfstudioDPConfig,GSTemplateModelConfig,GSTemplateModel,
    SplatfactoModel,AdamOptimizerConfig,ViewerConfig,
    MethodSpecification,ExponentialDecayScheduler,
    populate__modules,split__gaussians,compute__custom__losses,
    get__outputs,get__gaussian__param__groups,get__param__groups,
    get__outputs__for__camera,load__state__dict,get__metrics__dict,
    DepthDataManager,means,features__dc,features__rest,
    opacities,scales,quats,camera__opt,
    pyproject.toml,configs.py,model.py,losses.py,rendering.py,
    datamanagers.py,methods,
    xys,radii,gauss__params,batch,info},
  morecomment=[l]{//},
  frame=single,
  rulecolor=\color{cfg-frame},
  framesep=5pt,
  xleftmargin=5pt,
  xrightmargin=3pt,
  aboveskip=8pt,
  belowskip=5pt,
  breaklines=true,
  columns=fullflexible,
  keepspaces=true,
  showstringspaces=false,
  tabsize=4,
  literate=
    {::=}{{{\color{cfg-op}\bfseries::=}}}{3}
    {|}{{{\color{cfg-op}\bfseries\textbar}}}{1}
    {*}{{{\color{cfg-op}\bfseries*}}}{1}
    {+}{{{\color{cfg-op}\bfseries+}}}{1}
    {<}{{{\color{cfg-nt}$\langle$}}}{1}
    {>}{{{\color{cfg-nt}$\rangle$}}}{1}
    ,
}

\lstdefinestyle{cfgstyle-fig}{
  style=cfgstyle,
  basicstyle=\ttfamily\fontsize{6.8pt}{8.6pt}\selectfont,
  framesep=3.5pt,
  xleftmargin=3pt,
  xrightmargin=2pt,
  aboveskip=4pt,
  belowskip=2pt,
}
 \titlespacing*{\section}{0pt}{0pt}{0pt}
\titlespacing*{\paragraph}{0pt}{0pt}{1em}
\title{SplatLab: Autonomous Scientific Discovery for 3DGS Research}

\author{%
  Seemandhar Jain \qquad Manmohan Chandraker \\
  University of California San Diego \\
  \texttt{\{sejain,mkchandraker\}@ucsd.edu}
}

\newcommand{\method}{SplatLab}
\newcommand{\splatify}{\textsc{splatify}}
\newcommand{\nerfify}{\textsc{nerfify}}

\begin{document}

\maketitle

\begin{figure*}[h]
    \centering
    \includegraphics[width=\linewidth]{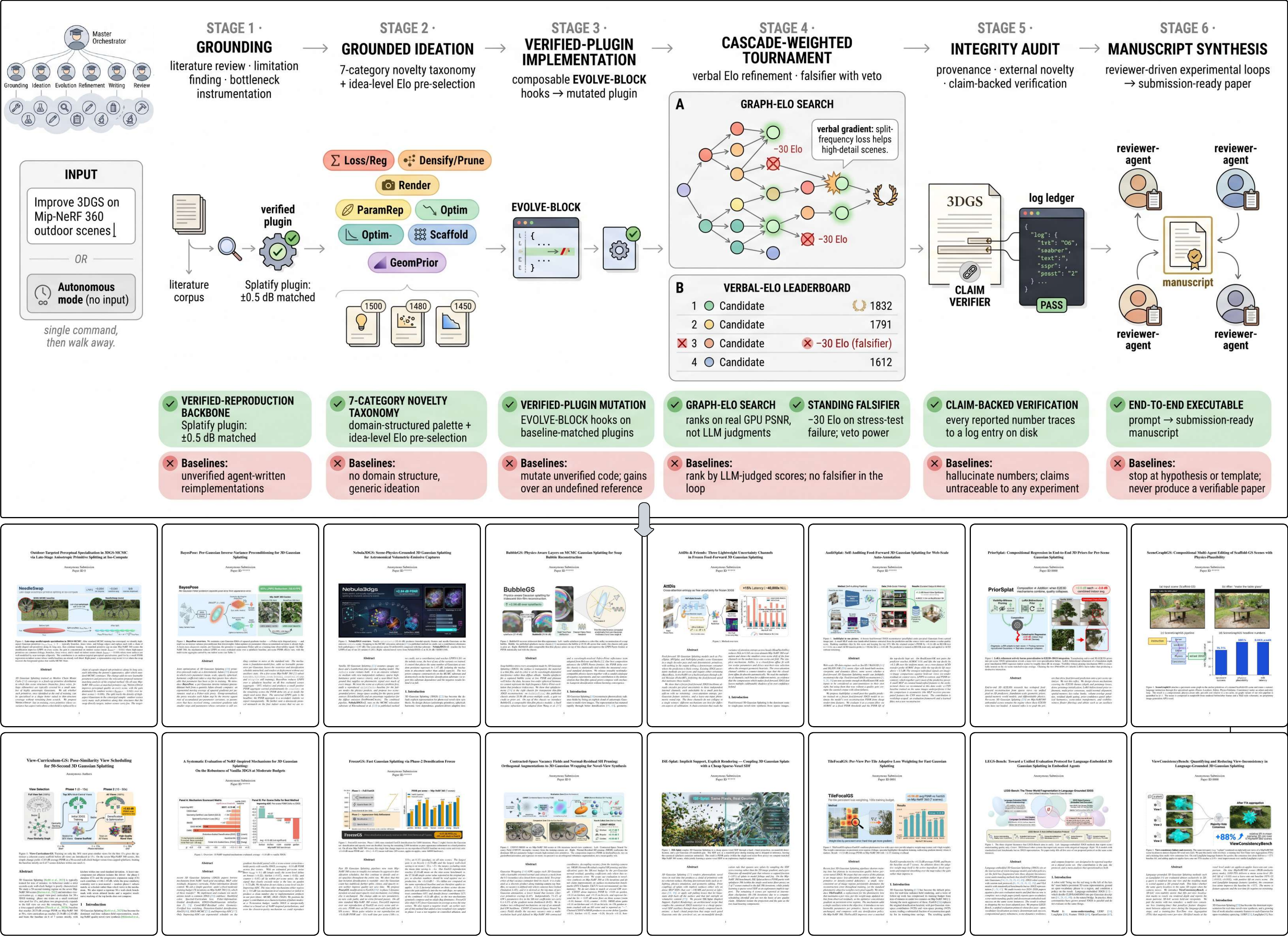}
\caption{Overview of \method{}: research pipeline (top) and sixteen generated manuscripts (bottom).}   \label{fig:teaser}
\end{figure*}

\begin{abstract}
Scientific discovery requires more than generating hypotheses: ideas must be grounded in prior work, translated into executable methods, experimentally validated, challenged by counterevidence, and communicated through reproducible findings. Despite recent advances in autonomous research agents, closing this entire loop remains challenging in rapidly evolving computer-vision fields with complex training pipelines and expensive evaluations.

We introduce \method{}, an autonomous scientific discovery system for 3D Gaussian Splatting (3DGS) that integrates literature understanding, hypothesis generation, implementation, experimental refinement, falsification, and manuscript writing into a unified research pipeline. Given a research prompt, a target-paper URL, or no input at all, \method{} identifies research opportunities, proposes novel methods, trains and ablates candidates across Mip-NeRF~360 and LLFF, and produces manuscripts grounded in verifiable experimental evidence.

Unlike prior systems, \method{} performs discovery over verified baseline implementations using domain-aligned modification scaffolds. A cascade-weighted Elo tournament evolves candidate methods using measured GPU performance and verbal-gradient feedback, while a standing falsifier incorporates adversarial stress testing directly into selection. An integrity audit binds reported claims to logged experiments, and reviewer-driven feedback can initiate further experimentation rather than merely revising text.

\method{} discovers methods achieving PSNR improvements of $+0.18$ to $+3.78$\,dB over matched baselines, including $+3.78$\,dB on astrophotographic nebula rendering, a previously unexplored setting for 3DGS. By closing the loop between scientific reasoning, executable experimentation, and evidence-backed communication, \method{} compresses months of researcher effort into days of mostly autonomous compute. Code, manuscripts, and experiment ledgers are available at \url{https://seemandhar.github.io/splatlab}.
\end{abstract}
\section{Introduction}
\label{sec:intro}

Automating scientific discovery is a long-standing research ambition,
traceable through early symbolic systems such as the Automated
Mathematician~\cite{wiki:automated_mathematician} and
DENDRAL~\cite{buchanan1981dendral}, the conceptual programme of
\emph{AI-generating algorithms}~\cite{clune2019ai}, and a contemporary line
of work in which LLMs compose research ideas, write
training code, run experiments, and produce
manuscripts~\cite{lu2024aiscientist,yamada2025ai,schmidgall2025agentlab,gottweis2025towards,liu2024aigs,tang2025airesearcher,feng2026internagent,song2026paperorchestra}.
The classical research workflow is bounded by an individual researcher's
time and reading bandwidth, neither of which scales with the publication
rate of an active subfield. Automating parts of the cycle could return
researcher time to ideation and analysis. Computer vision is a natural
place to test this: a single subfield now produces thousands of follow-up
papers in a few years, and reproducing one published method often takes
engineering effort comparable to producing a new one.

We are not aware of an autonomous-research pipeline demonstrated end-to-end in a computer-vision subfield where reproducing a single prior paper takes weeks of expert engineering. Existing systems edit short LaTeX templates over brief training runs, rank free-text hypotheses without executing a training loop, or build experimentation on top of LLM-written reimplementations that have not been externally verified against the target paper~\cite{lu2024aiscientist,yamada2025ai,gottweis2025towards,liu2024aigs,schmidgall2025agentlab,tang2025airesearcher,feng2026internagent,yuan2025dolphin,kon2025curie,jansen2025codescientist,li2024mlrcopilot,weng2024cycleresearcher}. PaperBench measures this verification gap: across 20 ICML~2024 Spotlight and Oral papers, the strongest evaluated agent reaches 21.0\% average replication~\cite{starace2025paperbench}. Without a verified reimplementation, an improvement claim has no fixed external reference: reported metrics may not trace to a measurement on disk, long-running agents drift from their objective as context fills, and comparisons are sometimes made against unspecified baselines. Table~\ref{tab:capability-matrix} summarises four common gaps: domain framework, verified baseline, full-loop refinement, and claim-to-log binding.

Progress on this problem appears to require a domain-specialised research
stack constructed on top of verified reproductions, rather than only
improvements in generalist coding ability. Generalist agentic
frameworks~\cite{qian2024chatdev,hong2024metagpt,wu2024autogen,yang2024sweagent,wang2024openhands}
were designed for software engineering, where every task starts from a
fresh codebase, not for the iterative refinement atop a verified baseline
that scientific research typically requires. Four design choices follow:
choose a single live subfield, build a verified-reproduction backbone for
it, expose a structured modification space to the ideator, and bind every
reported number to a logged GPU run.

We instantiate this for 3D Gaussian Splatting
(3DGS)~\cite{kerbl20233dgs}, an active subfield of neural rendering since
2023. Reproducing a 3DGS paper on top of \texttt{gsplat}~\cite{ye2025gsplat}
requires correct densification scheduling, opacity reset, spherical-harmonic
activation, and view-dependent appearance, none of which is supplied by the
prompt a generalist code agent receives; on a 30-paper 3DGS reproduction
benchmark, current frontier models succeed on roughly 5\% of
inputs~\cite{jain2026nerfify}. The verified-reproduction backbone has two
parts: \textsc{Nerfify}~\cite{jain2026nerfify} converts NeRF papers into
Nerfstudio plugins matching expert-written code; \textsc{Splatify}~\cite{jain2026splatifyreproducediscoverinnovate} extends the construction to 3DGS through a context-free grammar over a modular template
with nine composable extension hooks. Every candidate that reaches our
experimentation phase mutates a \textsc{Splatify} plugin already verified
to train, and match its target.

We present \method{}, an execution-grounded research system demonstrated in 3D Gaussian Splatting that produces research manuscripts with verified experiments from a one-line prompt, a target-paper URL, or no input at all, with every reported number traced to a logged GPU run. \method{} orchestrates 36 agents across seven phases (research, ideation, implementation, experimentation, analysis, writing, peer review) without a fixed DAG. Five design decisions distinguish \method{} from prior autonomous-research and program-search systems; each is detailed in Section~\ref{sec:method}. \textit{(i) Grounded Elo on real GPU PSNR.} Idea ranking uses cascade-weighted, grounding-conditional Elo scored on measured PSNR rather than the model-judged proposal scores of AI Co-Scientist~\cite{gottweis2025towards}, with ReEvo-style verbal gradients~\cite{ye2024reevo} fed back to the ideator. \textit{(ii) In-loop falsifier coupled to selection.} A standing adversarial falsifier extending AIGS~\cite{liu2024aigs} runs in-loop; each failed attack deducts Elo, making stress-test survival a selection signal rather than a post-hoc check. \textit{(iii) Claim-backed verification.} Every number that reaches a draft is bound to a logged experiment, and the run cannot terminate while any claim lacks a backing measurement, a binding absent from prior pipelines~\cite{lu2024aiscientist,yamada2025ai}. \textit{(iv) Verified-reproduction backbone.} Every candidate mutates a \textsc{Splatify} plugin already verified against its target, providing a fixed reference for every reported gain. \textit{(v) Domain-aligned modification space.} Seven 3DGS novelty categories wrap EVOLVE-BLOCK markers and the \splatify{} island tournament~\cite{jain2026splatifyreproducediscoverinnovate} in category-aligned scaffolds; a cross-codebase auto-mapper places them in arbitrary 3DGS repositories so the diff space coincides with the field's modification space.

On Mip-NeRF~360~\cite{barron2022mipnerf360} and
LLFF~\cite{mildenhall2019llff}, \method{} produces manuscripts that
each beat their reproduced target by $+0.18$ to $+3.78$\,dB PSNR,
including $+3.78$\,dB on astrophotographic nebula rendering, a domain not
addressed by prior 3DGS work. The research cycle (read, reproduce, extend,
train, ablate, write) compresses from months of graduate-student time to
days of mostly autonomous compute, enabling design-space sweeps not
feasible at single-researcher pace. \method{}, all agents, and the code can be found at \url{http://seemandhar.github.io/splatlab}.
\section{Related Work}
\label{sec:related}

\paragraph{End-to-end automated research pipelines.}
\method{} is closest to goal-to-manuscript systems. The AI Scientist and v2 produce ML papers via fixed templates and agentic tree search~\cite{lu2024aiscientist,yamada2025ai}. Agent Laboratory, MLR-Copilot, AI-Researcher, NovelSeek/InternAgent, Dolphin, and CodeScientist organize the literature-to-paper loop with role-specialized agents~\cite{schmidgall2025agentlab,li2024mlrcopilot,tang2025airesearcher,zhang2025novelseek,yuan2025dolphin,jansen2025codescientist}; CycleResearcher post-trains an LLM on a research-review-refine cycle, while AgentRxiv shares a preprint pool across runs~\cite{weng2024cycleresearcher,schmidgall2025agentrxiv}. AIGS adds a post-hoc \textit{FalsificationAgent}, data-to-paper traces claims to data, Curie embeds rigor via agent control modules, and PaperBench grades replication against author rubrics~\cite{liu2024aigs,ifargan2025datatopaper,kon2025curie,starace2025paperbench}. All rely on agent-written reimplementations unverified against their targets; \method{} requires every candidate to mutate a plugin verified to compile, train, and match its target's headline. None condition on $(B,L)$, fuse falsification with cascade-weighted Elo, or allow reviewer-spawned re-experimentation rather than only prose edits.

\paragraph{Autonomous discovery without manuscript output.}
Other systems run the scientific loop but stop short of a paper. Google's AI co-scientist uses a multi-agent Elo tournament over 203 biomedical goals with wet-lab validation in drug repurposing and antimicrobial resistance~\cite{gottweis2025towards}. The Virtual Lab designed 92 experimentally validated SARS-CoV-2 nanobodies~\cite{swanson2025virtuallab}; Coscientist and ChemCrow connect GPT-4 to wet-lab and computational chemistry tools~\cite{boiko2023coscientist,bran2024chemcrow}. ResearchAgent and Si et al.\ study LLM ideation in isolation; the latter's 100+ NLP-researcher study found LLM ideas judged more novel but slightly less feasible than expert ideas~\cite{baek2025researchagent,si2024can}. Wei et al.\ survey the landscape~\cite{wei2025agenticscience}; we instantiate their Level-3 ``full agentic discovery'' in graphics rather than wet-lab biology. Closest to our tournament is AI co-scientist's Elo loop, but it ranks free-text hypotheses rather than compiled training runs against a $(B,L)$ substrate, and updates ratings uniformly rather than by evaluation depth.

\paragraph{Evolutionary program search and multi-agent infrastructure.}
\method{}'s island tournament with EVOLVE-BLOCK markers builds on LLM-driven program evolution. FunSearch pairs an LLM with a systematic evaluator in an island loop~\cite{romera2024funsearch}; AlphaEvolve generalizes this to entire codebases via a Gemini ensemble~\cite{novikov2025alphaevolve}. EoH co-evolves natural-language ``thoughts'' and code; ReEvo introduces \textit{verbal gradients} (LLM reflections on why one heuristic beat another, fed back into mutation); Promptbreeder evolves prompts self-referentially; Mind Evolution scales inference-time compute through genetic search~\cite{liu2024eoh,ye2024reevo,fernando2024promptbreeder,lee2025mindevolution}. We route ReEvo's verbal gradients to the ideator rather than a heuristic mutator, and replace per-individual fitness with cascade-weighted, grounding-conditional Elo so cheap dry-run wins do not dominate full-train candidates against the verified substrate. MetaGPT, ChatDev, AutoGen, and SWE-agent supply the orchestration primitives we reuse~\cite{hong2024metagpt,qian2024chatdev,wu2024autogen,yang2024sweagent}, while LLM-as-a-judge, MARG, and AgentReview inform our reviewer-driven re-experimentation but treat review as terminal rather than a gate demanding new substrate experiments~\cite{zheng2023llmjudge,darcy2024marg,jin2024agentreview}.

\section{Method}
\label{sec:method}

\begin{figure*}[t]
    \centering
    \includegraphics[width=\linewidth]{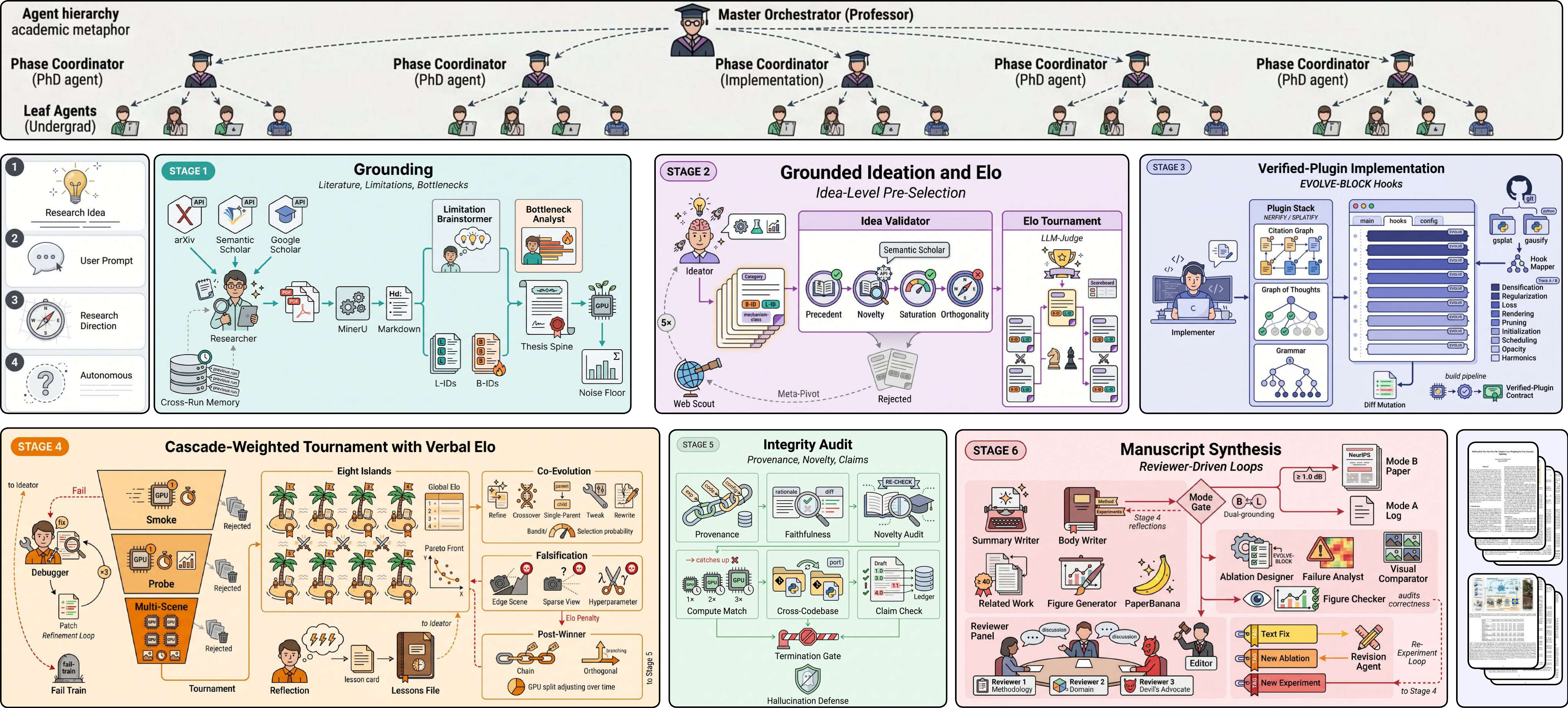}
    \caption{\textbf{Detailed pipeline of \method{}.} Expanding the overview in Figure~\ref{fig:teaser}, a research intent (or no input) passes through six stages (grounding, Elo-pre-ranked ideation, \texttt{EVOLVE-BLOCK} implementation, cascade-weighted Elo refinement across eight islands, integrity audit, and reviewer-driven manuscript synthesis) run by a three-level hierarchy of Professor,
    PhD Coordinators, and Undergrad Leaf Agents.}
    \label{fig:pipeline}
\end{figure*}

\method{} is a system for \emph{verified evolutionary research over a reproducible substrate}. The substrate is a 3DGS reproduction backbone (\splatify{}~\cite{jain2026splatifyreproducediscoverinnovate}/\nerfify{}~\cite{jain2026nerfify}) whose plugins match their published targets; every extension \method{} proposes mutates a plugin already verified to match its target. This scaffolding separates \method{} from prior systems that pursue unconstrained autonomous discovery over unverified agent-written code, whose drafts are neither traceable to logs nor reproducible by retraining. The claim is narrower and stronger: \method{} performs \emph{autonomous iterative discovery within a live, technically demanding subfield}, with every iteration anchored to a measured baseline and bound to a logged training run.

Formally, given a research prompt, a target-paper URL, or no input at all, \method{} returns $P = (P_{\text{tex}},\, P_{\text{pdf}},\, \mathcal{L},\, \mathcal{C}) = W(x;\, \mathcal{D})$, where $P_{\text{tex}}$ is the LaTeX source, $P_{\text{pdf}}$ the compiled PDF, $\mathcal{L}$ binds every number in $P_{\text{tex}}$ to a training run on disk, and $\mathcal{C}$ is the codebase that produced those numbers. The fixed context $\mathcal{D} = (\mathcal{R}, \mathcal{B}, \mathcal{H}, \mathcal{G})$ supplies the 3DGS literature corpus $\mathcal{R}$, the \splatify{} backbone $\mathcal{B}$, the nine \texttt{EVOLVE-BLOCK} hooks $\mathcal{H}$, and the target conference template $\mathcal{G}$. \method{} produces both $\mathcal{C}$ and $\mathcal{L}$ in the same run against the verified $\mathcal{B}$; its writing stage can re-enter implementation to spawn new training runs.

Fig.~\ref{fig:pipeline} decomposes $W$ into six stages, each acting on the same verified substrate: \emph{Grounding} instruments a measured bottleneck on the matched plugin; \emph{Ideation} pre-ranks candidate fixes by idea-level Elo; \emph{Implementation} compiles survivors into \texttt{EVOLVE-BLOCK} mutations of the plugin; \emph{Refinement} runs a cascade-weighted Elo tournament across eight islands with standing falsification, scored on real \texttt{ns-train} PSNR; \emph{Audit} verifies that every reported number traces back to a measurement on the verified substrate; \emph{Synthesis} writes the manuscript through reviewer-driven loops that may re-enter Stages 3--4. Stages run under a three-level hierarchy (Professor, PhD coordinators, Undergrad leaf agents; 36 total) with no fixed DAG: the master may revisit any earlier stage on a falsifier veto, audit failure, or reviewer-spawned ablation.

\subsection{\method{}}
\label{sec:pipeline}

\paragraph{Stage 1: Grounding.}
Stage 1 anchors the run in the substrate. Two parallel investigations produce the \emph{thesis spine} $\Gamma$, the only artefact Stages 2--6 condition on. The first reads the literature: a Researcher queries arXiv, Semantic Scholar, and Google Scholar and pulls from a persistent cross-run paper store via RAG, with retrievals written back before Stage 1 closes so the corpus grows monotonically across runs. Retrieved PDFs are parsed by MinerU~\cite{wang2024mineru} and a Limitation Brainstormer cross-references each paper's stated limitations against critiques in later work, emitting \textbf{L-IDs}: deduplicated limitation strings such as \texttt{L-127: opacity reset destabilises late refinement}, each with its supporting citation set. The second reads the substrate: a Bottleneck Analyst trains the verified \splatify{} plugin on representative Mip-NeRF~360 scenes and profiles per-tile loss, PSNR trajectory, memory, and gradient norms; patterns become \textbf{B-IDs}: measurement-anchored strings such as \texttt{B-04: PSNR plateau at $25.65$\,dB on garden}, each tagged with the \texttt{EVOLVE-BLOCK} category it implicates. $\Gamma$ contains the \splatify{} reproduction vector $\mathcal{V}$, the L-IDs, and the B-IDs.

\paragraph{Stage 2: Grounded ideation.}
Stage 2 turns $\Gamma$ into a ranked set of candidates worth running on the substrate, all without running any training. A candidate is a tuple $(\texttt{mechanism}, c, l, b, \hat{\Delta})$, where $c \in \{1,\ldots,7,\,\texttt{open}\}$ indexes one of seven 3DGS novelty categories, $l$ is the L-ID it claims to address, $b$ the B-ID it claims to attack, and $\hat{\Delta}$ a predicted delta on the headline metric. An Ideator generates candidates conditioned on $\Gamma$ over five iterations; between rounds, a Web Scout fires fresh literature searches keyed on the previous round's L-IDs and B-IDs, so prior art widens monotonically rather than re-sampling from a fixed window~\cite{si2024can}. Each candidate clears four checks: \emph{Precedent} (the L-ID/B-ID citation set supports the claim), 
\emph{Novelty} (neither a restatement of prior work
nor a parameter-only tweak lacking a new signal,
state, or mechanism, as screened by the
\emph{Tweak Trap}), \emph{Saturation} (fewer than $T$ logged negatives in the category-mechanism cell), and \emph{Orthogonality} (not a near-duplicate of another admitted candidate); rejection routes back to the Web Scout as a \emph{Meta-Pivot} signal. Survivors enter a bracket tournament with initial Elo $1200$, judged by an LLM-as-Judge along four axes: novelty against the L-ID citation set, specificity to the B-ID, implementability within the category, and predicted-delta plausibility. Top-quartile uses multi-turn debate; lower tiers use single-turn pairwise comparisons. The output passes into Stage 3 where each candidate becomes a real plugin mutation, and into the GPU-grounded tournament in Stage 4.

\paragraph{Stage 3: Verified-plugin implementation.}
Stage 3 mutates the substrate. Each surviving idea becomes a code change to a \splatify{}/\nerfify{} plugin~\cite{jain2026splatifyreproducediscoverinnovate,jain2026nerfify} that already compiles, trains, and matches the target's published headline; the verified-plugin contract attributes downstream gains to the proposed mechanism rather than to silent reimplementation drift. A Hook Mapper resolves the abstract category $c$ into a concrete subset of the nine \texttt{EVOLVE-BLOCK} hooks (densification, regularisation, loss, rendering, pruning, initialisation, scheduling, opacity, harmonics) and places \texttt{EVOLVE-BLOCK} markers around the corresponding regions, following~\cite{novikov2025alphaevolve} but over a verified plugin rather than agent-written code; for $c = \texttt{open}$, a cross-codebase auto-mapper locates insertion points that span existing hooks, compose them, or introduce new ones. 
\texttt{EVOLVE-BLOCK}s are paired code comments that delimit regions agents can modify while preserving the surrounding
verified implementation.
An Implementer generates the code under three constraints: a \emph{Citation Graph} retrieves L-ID/B-ID-anchored papers and their published implementations as reference code; a \emph{Graph of Thoughts}~\cite{besta2024got} controller structures reasoning into generation, aggregation, and refinement nodes so a failed branch can be revisited rather than restarted; and a context-free \emph{Grammar} of \splatify{} restricts output to legal hook-level rewrites. Mutations are emitted as structured diffs~\cite{novikov2025alphaevolve} rather than full rewrites. The diff is applied, the plugin compiled, and a $100$-iteration smoke test trains on a single Mip-NeRF~360 scene; a candidate that compiles, finishes the smoke test stably, and instantiates the requested hooks clears the contract. Failures return to the Implementer with the build log as a verbal-gradient signal, capped at a repair budget before logging to the cross-run store as a build-failure negative.

\paragraph{Stage 4: Cascade-weighted tournament.}
Stage 4 makes the substrate the judge. Each candidate is trained and scored on real PSNR; ratings are conditioned on the grounding pair $(B, L)$ (benchmark ID, loss ID), so strength is tracked per evaluation context rather than as a scalar leaderboard.

Each candidate flows through four stages of increasing rigour: \emph{smoke} ($100$-iteration sanity check), \emph{probe} (full-iteration on one scene), \emph{multi-scene} (full-iteration on a stratified subset), and the full benchmark, which only the champion enters. Stage $s$ updates Elo with weight $w_s \in \{0.2, 0.5, 1.0\}$ from smoke through multi-scene, so a smoke win cannot drown out a multi-scene loss; without this gating, a three-generation eight-island run on Mip-NeRF~360 would exceed multiple GPU-months. For a match between candidates $i$ and $j$ at stage $s$,
\begin{equation}
    R_i^{(B,L)} \leftarrow R_i^{(B,L)} + K \cdot w_s \cdot \bigl( S_i - E_i^{(B,L)} \bigr), \qquad
    E_i^{(B,L)} = \frac{1}{1 + 10^{(R_j^{(B,L)} - R_i^{(B,L)})/400}},
    \label{eq:elo-update}
\end{equation}
with $K = 32$, initial rating $1200$, and $S_i \in \{0, 0.5, 1\}$ from the headline-metric comparison. Each update is paired with an LLM rationale naming the regime, metric delta, and likely causal mechanism; rationales accumulate in a \emph{Lessons File} fed back to the Ideator as a verbal gradient~\cite{ye2024reevo}. Smoke failures route to a Debugger for up to three patches before logging to the \emph{Fail Train}.

Survivors are distributed across eight islands and mutated by five operators selected by a multi-armed bandit over success rates: \emph{Refine} (local rewrites), \emph{Crossover} (two-parent mixing), \emph{Single-Parent} (one-parent mutation), \emph{Tweak} (hyperparameter perturbation), and \emph{Rewrite} (full-block regeneration). Per-island Elo distributions form a Pareto front over $(B, L)$; promotion requires Pareto improvement, not scalar dominance. A standing falsifier attacks every survivor along three axes: edge scenes (OOD trajectories, extreme exposure), sparse-view reduction, and hyperparameter perturbation within published ranges. Each failed attack subtracts $\delta = 30$ Elo, making stress-test survival a selection signal rather than a post-hoc check; the falsifier holds veto power before manuscript writing. Once a champion emerges, the remaining GPU budget splits between \emph{chain} (sequential variants extending the winner) and \emph{orthogonal} (branches recombining components from non-champion islands), shifting from chain-heavy to orthogonal-heavy as the champion's per-iteration gain saturates.

\paragraph{Stage 5: Integrity audit.}
Stage 5 binds the draft back to the substrate. Six audits run between the Stage 4 champion and the draft; any failure aborts the run as a research log. \emph{Provenance}: a content-hash chain links every reported number to its (experiment ID, commit, config) triple. \emph{Faithfulness}: the Stage 3 rationale is checked against the diff actually applied, catching drift that PaperOrchestra~\cite{song2026paperorchestra} cannot reach by construction. \emph{Novelty}: the mechanism is re-resolved against Semantic Scholar with a query expanded by post-hoc results, catching prior work the Stage 2 Validator missed. \emph{Compute match}: the headline metric is recomputed under the baseline's compute envelope; gains that fail are reclassified as compute-confounded. \emph{Cross-codebase port}: the winning hook is ported via the Stage 3 auto-mapper to a second 3DGS codebase indexed by \splatify{}; gains that fail are flagged as codebase-specific. \emph{Claim check}: every numerical or theoretical claim is resolved against the experiment ledger; load-bearing mismatches terminate the run, closing the hallucination failure mode that AI~Scientist~\cite{lu2024aiscientist} drafts leave open.

\paragraph{Stage 6: Manuscript synthesis.}
Stage 6 turns substrate-grounded results into a manuscript that, when challenged, re-enters the substrate. The $(B, L)$ pair routes the run: a new discovery or beating SOTA triggers Mode B (manuscript synthesis), otherwise Mode A (research log only). Five writers run in parallel: Summary Writer (abstract, introduction), Body Writer (method, experiments; consumes the Stage 4 Lessons File), Related Work (citation floor enforced via the Stage 1 Researcher), Figure Generator (result plots from the ledger), and PaperBanana~\cite{zhu2026paperbanana} (conceptual diagrams). A VLM Figure Checker audits axis labels, legend completeness, and number-vs-caption agreement; an Ablation Designer reads the \texttt{EVOLVE-BLOCK} diff and emits leave-one-out ablations, a Failure Analyst surfaces residual-error heatmaps, and a Visual Comparator tiles qualitative renders against the reproduced baseline. Three reviewer agents argue under an Editor (Methodology, Domain, Devil's Advocate); the panel extends AgentReview~\cite{jin2024agentreview} by demanding new training runs, not only prose edits. New experiments can re-enter Stage 4 with the comment as a verbal-gradient seed. The loop terminates when no reviewer raises a New-Experiment concern.

\section{Experiments}
\label{sec:experiments}

We evaluate two aspects of \method{}: the research process, through run outcomes, component ablations, failure analysis, and compute costs; and its outputs, through method performance, idea and manuscript quality, and human and multi-agent peer review. Detailed run outcomes and failure cases are reported in the appendix.

\begin{figure*}[h]
    \centering
    \includegraphics[width=\linewidth]{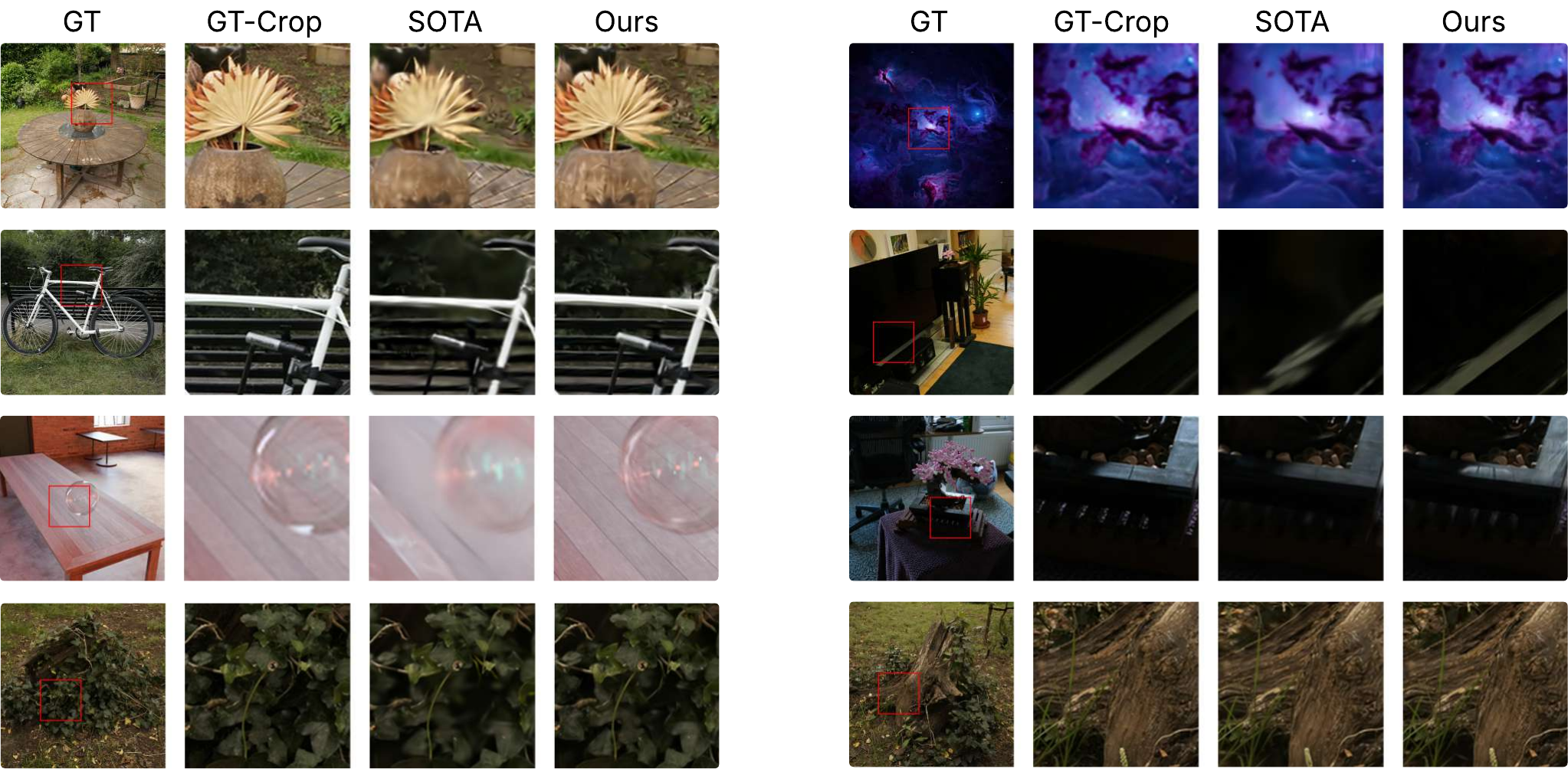}
\caption{\textbf{Qualitative comparison against per-regime SOTA.} Columns: ground truth with evaluated region boxed (\textbf{GT}), zoomed crop (\textbf{GT-Crop}), the leading published 3DGS method for that row's regime (\textbf{SOTA}, varies per row; see Tab.~\ref{tab:quant}), and the \method{}-discovered method (\textbf{Ours}).}    \label{fig:qual}
\end{figure*}

\begin{table}[h]
\centering\footnotesize
\caption{\method\ vs.\ SOTA on a representative subset of target prompts. Top: novel scenes with vanilla 3DGS as baseline. Bottom: SOTA improvements on standard 3DGS benchmarks. \emph{Limitation/fix}: analyst's diagnosis (SOTA row) and validated mechanism (\emph{ours} row); bold marks the winner in each pair. The full table covering the $+0.18$ to $+3.78$\,dB PSNR range is in the Appendix.}\label{tab:quant}
\setlength{\tabcolsep}{2.5pt}
\renewcommand{\arraystretch}{0.85}
\resizebox{\textwidth}{!}{%
\begin{tabular}{@{}ll>{\scriptsize}p{6.4cm}rrr@{}}\toprule
Setting & Method & \normalsize Limitation attacked / fix & PSNR$\uparrow$ & SSIM$\uparrow$ & LPIPS$\downarrow$ \\
\midrule
\multicolumn{6}{@{}l}{\textbf{Novel scenes} (vanilla 3DGS baseline)} \\
\midrule
\multirow{2}{*}{Nebula}
  & 3DGS~\citep{kerbl20233dgs}              & pinhole ignores OPENCV distortion; SfM gives $\approx\!0$ stars  & 32.58 & 0.914 & 0.278 \\
  & \textbf{Nebula3DGS-v2} \emph{(ours)}    & 3DGS-MCMC sampler with saturation-driven star seeder             & \textbf{36.36} & \textbf{0.945} & \textbf{0.201} \\
\midrule
\multirow{2}{*}{Bubble}
  & 3DGS~\citep{kerbl20233dgs}              & SH can't represent thin-film iridescence; densification stalls on emissive translucency & 27.40 & 0.857 & 0.337 \\
  & \textbf{BubbleGS} \emph{(ours)}         & 3DGS-MCMC chassis, Snell/Fresnel two-pass conditional shell renderer & \textbf{30.11} & \textbf{0.906} & \textbf{0.224} \\
\midrule
\multicolumn{6}{@{}l}{\textbf{Improving published SOTA}} \\
\midrule
\multirow{2}{*}{MipNeRF360}
  & FastGS~\citep{ren2026fastgs}            & $\{0,1\}$ thresholded pixel-count is frequency-blind             & 29.12 & 0.858 & 0.217 \\
  & \textbf{TileFocalGS} \emph{(ours)}      & rank-graded per-tile weights, EMA-smoothed across views          & \textbf{29.69} & \textbf{0.869} & \textbf{0.194} \\
\midrule
\multirow{2}{*}{MipNeRF360}
  & splatfacto~\citep{tancik2023nerfstudio} & @50\,s wall: severely under-converged; densification not stabilised  & 20.53 & 0.593 & 0.505 \\
  & \textbf{View-Curriculum-GS} \emph{(ours)} & first 15\,s on 30\% most pose-similar views, then full set     & \textbf{23.36} & \textbf{0.686} & \textbf{0.412} \\
\midrule
\multirow{2}{*}{MipNeRF360}
  & MSAA-3DGS~\citep{yan2024msaa3dgs}       & multi-sample AA blurs fine contrast; underconverged at 30K iters & 25.35 & 0.758 & 0.369 \\
  & \textbf{MSAA-Bilat} \emph{(ours)}       & bilateral grid contrast recovery, $+15$K extra iters             & \textbf{26.19} & \textbf{0.791} & \textbf{0.293} \\
\midrule
\multirow{2}{*}{LLFF}
  & DWTGS~\citep{nguyen2025dwtgs}           & wavelet basis under-constrained without high-freq supervision    & 18.68 & 0.510 & 0.374 \\
  & \textbf{DWTGS-Patch} \emph{(ours)}      & patch structure loss, opacity/normal/depth regularisers          & \textbf{21.09} & \textbf{0.725} & \textbf{0.281} \\
\bottomrule
\end{tabular}}
\end{table}

\paragraph{Experimental Setup. }
\label{sec:setup}

We run \method{} multiple times: a mix of \emph{prompted} runs ($x$
is a target-paper SOTA), single prompt and \emph{autonomous} runs ($x = \varnothing$,
the system selects its own target from the literature);
per-run mode and target are listed in supp. None of the runs required manual code edits after dispatch. Each run's \splatify{}
plugin or original source code is verified against its target's published headline or, against the paper reported numbers, before \method{} begins
(Table~\ref{tab:quant}). Standard benchmarks are
Mip-NeRF~360~\cite{barron2022mipnerf360} (nine unbounded scenes,
including \textit{flowers} and \textit{treehill}) and
LLFF~\cite{mildenhall2019llff}; every
8th image is held out for testing
following~\cite{bagdasarian20253dgszip}, with all training runs at
30k iterations on NVIDIA A6000 GPUs. We test inter-domain
discovery on three out-of-distribution scenes captured for this
work and not previously addressed by 3DGS: \textit{nebula}
(astrophotographic), \textit{bubble}, and \textit{diamond}. We
report PSNR (dB)$\uparrow$, SSIM$\uparrow$, and
LPIPS$\downarrow$. 

Our experiments are designed to demonstrate not equalized search budgets, but system capability under end-to-end operation (with long-run autonomy a critical challenge that we overcome). Besides scaling, our benefits accrue through the combination of verified reproduction, structured mutation space and grounded evaluation. Reproduction details, the LLM stack, per-run compute budgets, and per-scene breakdowns are in the supplementary.

\paragraph{Run outcomes.}
Across 16 runs, \method{} trained 198 experiment branches, yielding five verified positives, five marginal results with qualified claims, five non-PSNR contributions, and one recorded failure. The Appendix details selection rejections, training failures, and outcomes by target regime.

\paragraph{Compute cost.}
Runs used a 24--48-hour wall-clock budget, averaging 113M API tokens across approximately 1,000--1,500 LLM calls. GPU consumption ranged from 2--36 GPU-hours per run, with estimated LLM API costs of \$150--250. Detailed accounting appears in the Appendix.

\paragraph{Generated Methods vs.\ SOTA.}
\label{sec:sota}

We evaluate \method{} on multiple targets in two regimes. Three are out-of-distribution scenes (Nebula, Bubble, Diamond) with vanilla 3DGS~\citep{kerbl20233dgs} as the baseline; the remaining eight target published 3DGS methods on Mip-NeRF~360 and LLFF, where the discovered variant must beat a tuned, peer-reviewed reference under the original paper's iteration count, scene set, and compute budget. Table~\ref{tab:quant} pairs each baseline with the \method{}-discovered method on PSNR, SSIM, and LPIPS; Figure~\ref{fig:qual} shows the regime-matched qualitative comparison. The discovered method exceeds its baseline on PSNR in every pairing, with gains from $+0.18$ to $+3.78$\,dB, the largest on the astrophotographic nebula scene, a regime where 3DGS has not previously been applied.

\paragraph{Multi-seed Robustness.}
To evaluate seed sensitivity, we reran GNS-AA with five seeds.
GNS achieved $25.95\pm0.07$ dB, while GNS-AA achieved
$26.18\pm0.05$ dB, yielding $+0.23\pm0.09$ dB compared
with the originally reported $+0.18$ dB. Other generated
studies also use three or four seeds. Improvements within
the estimated noise floor are reported as marginal rather
than verified gains.

\paragraph{\method{} vs.\ Prior Discovery Systems. }
\label{sec:vs-discovery}

Table~\ref{tab:capability-matrix} positions \method\ against the eight closest autonomous-research and discovery systems on the capability axes of Section~\ref{sec:method}. \method\ is the only system that satisfies all five rows anchoring our contributions: a verified-reproduction backbone for a live subfield, grounding-conditional idea re-ranking, a standing falsifier coupled to selection, claim-backed verification, and a reviewer panel that can spawn new training runs rather than only prose edits. Full results including wall clock and token comparison in the supplementary.

\newcommand{\cmark}{\ensuremath{\checkmark}}
\newcommand{\xmark}{\ensuremath{\text{\sffamily X}}}
\newcommand{\pmark}{\raisebox{0.15ex}{\scalebox{0.9}{$\sim$}}}


\begin{table}[h]
  \centering\footnotesize
  \caption{Capability comparison against the closest autonomous-research and program-search 
    systems. \cmark = supported, \pmark = partial, \xmark = not supported. Verdicts verified 
    against each cited paper; \pmark{} indicates a related but weaker mechanism (post-hoc 
    rather than in-loop, LLM-judged on outputs rather than measured metrics, manually placed 
    rather than auto-mapped, etc.). Full matrix (22 rows, grouped by stage, with per-cell justifications) in the Appendix.}
  \label{tab:capability-matrix}
  \resizebox{\linewidth}{!}{%
  \begin{tabular}{@{}lccccccccc@{}}
    \toprule
    Capability 
      & \makecell{AIS-v2\\\citep{yamada2025ai}} 
      & \makecell{A-Lab\\\citep{schmidgall2025agentlab}} 
      & \makecell{MLR\\\citep{li2024mlrcopilot}} 
      & \makecell{AIGS\\\citep{liu2024aigs}} 
      & \makecell{Curie\\\citep{kon2025curie}} 
      & \makecell{PaperOrch.\\\citep{song2026paperorchestra}} 
      & \makecell{CoSc\\\citep{gottweis2025towards}} 
      & \makecell{AlE\\\citep{novikov2025alphaevolve}} 
      & \textbf{Ours} \\
    \midrule
    End-to-end executable (no human in loop)                           & \cmark & \cmark & \pmark & \pmark & \cmark & \xmark & \pmark & \cmark & \cmark \\
    Produces a manuscript                                              & \cmark & \cmark & \xmark & \xmark & \xmark & \cmark & \xmark & \xmark & \cmark \\
    Real-experiment execution                                          & \cmark & \cmark & \cmark & \pmark & \cmark & \xmark & \xmark & \cmark & \cmark \\
    Live-subfield domain backbone                                      & \xmark & \xmark & \xmark & \xmark & \xmark & \xmark & \xmark & \xmark & \cmark \\
    Verified baseline reproduction                                     & \xmark & \xmark & \xmark & \xmark & \xmark & \xmark & \xmark & \xmark & \cmark \\
    Idea ranking on real GPU metrics                                   & \xmark & \pmark & \pmark & \xmark & \pmark & \xmark & \xmark & \cmark & \cmark \\
    Category-aligned scaffolds                                         & \xmark & \xmark & \xmark & \xmark & \xmark & \xmark & \xmark & \xmark & \cmark \\
    \midrule
    \multicolumn{10}{@{}l}{\emph{Verification and audit}} \\
    Reproducibility / rigor checks                                     & \xmark & \xmark & \xmark & \pmark & \cmark & \xmark & \xmark & \pmark & \cmark \\
    Claim-backed-verification                               & \xmark & \xmark & \xmark & \xmark & \xmark & \xmark & \xmark & \xmark & \cmark \\
    Simulated peer review                                              & \cmark & \cmark & \xmark & \xmark & \xmark & \cmark & \xmark & \xmark & \cmark \\
    Reviewer panel can spawn new training runs (not just edits)        & \xmark & \xmark & \xmark & \xmark & \xmark & \xmark & \xmark & \xmark & \cmark \\
    \bottomrule
  \end{tabular}%
  }
\end{table}

\paragraph{Idea and Manuscript Quality. }
\label{sec:quality}

\begin{table}[h]
  \centering\footnotesize
  \caption{Idea quality (blind expert protocol from~\citep{si2024can}). Forty
  style-normalised ideas (twenty from \method{} Stage~2, ten from 3DGS PhD students,
  ten from accepted preprints held out from the cross-run paper store) scored by
  ten experts on a 1--10 Likert scale.}
  \label{tab:idea-quality}
  \begin{tabular}{@{}lcccccc@{}}
    \toprule
    Source & $N$ & Novelty & Excitement & Feasibility & Effectiveness & Overall \\
    \midrule
    PhD students                 & 10 & $5.2$           & $5.0$           & $\mathbf{6.4}$ & $5.7$           & $5.1$ \\
    Accepted arXiv preprints     & 10 & $5.8$           & $5.5$           & $6.0$          & $5.8$           & $6.0$ \\
    \textbf{\method{} (Stage~2)} & 20 & $\mathbf{5.9}^{*}$ & $\mathbf{5.6}^{*}$ & $6.2$          & $\mathbf{6.1}^{*}$ & $\mathbf{5.8}^{**}$ \\
    \bottomrule
  \end{tabular}
\end{table}

We evaluate \method{} at two stages of its pipeline: ideation (Table~\ref{tab:idea-quality}) and final manuscript (Table~\ref{tab:manuscript-quality}). At ideation, twenty Stage~2 candidates are scored against ten PhD-student ideas and ten recently accepted workshop / conferences ideas under the blind-expert protocol of~\citep{si2024can}. At manuscript, three baseline pipelines each produce a draft from five human-verified prompts reverse-engineered from \method{} runs; drafts are scored against the published 3DGS paper for each prompt under the LLM-as-a-Judge paired protocol of \citep{song2026paperorchestra}, verified by graphics-area human reviewers. \method{} outperforms prior systems by large margins at both stages and scores close to human ground truth at the manuscript level.

\begin{table}[h]
  \centering\footnotesize
\caption{Manuscript quality on a 0--100 Multi-agent as judge~\cite{jin2024agentreview} (paired protocol from~\citep{song2026paperorchestra}); means over five human-verified prompts reverse-engineered from \method{} runs. \emph{Human (GT)}: the published 3DGS paper. \emph{SingleAgent}: \method{} with one frontier agent. Baselines share the same research idea; \emph{AI Scientist-v2}~\citep{yamada2025ai} and \emph{GPT-5 ET} receive Stage~2--3 materials, while PaperOrchestra additionally receives verified code, logs, and the Stage~5 audit summary. Default tool access is retained. Scores measure writing quality, not end-to-end discovery. Three graphics-area reviewers re-scored the ten manuscripts (five \method{},
five human) and audited the LLM judge's per-rubric rationales against the
manuscript text; reviewer scores agree with the LLM at Spearman
$\rho=0.87$ (weighted $\kappa=0.84$), and 54/60 of the LLM rationales were
marked as supported by manuscript content.}
\label{tab:manuscript-quality}
  \resizebox{\linewidth}{!}{%
  \begin{tabular}{@{}lccccccc@{}}
    \toprule
    Pipeline & 
      \makecell{Citation\\Practices} & 
      \makecell{Coverage \&\\Completeness} & 
      \makecell{Critical Analysis\\\& Synthesis} & 
      \makecell{Organisation\\\& Writing} & 
      \makecell{Positioning\\\& Novelty} & 
      \makecell{Relevance\\\& Focus} & 
      \makecell{Overall\\Score} \\
    \midrule
    \multicolumn{8}{c}{\emph{Judge: Gemini-3.1-Pro}} \\
    \midrule
    Human (GT)                 & 94 & 93 & 91 & 94 & 89 & 91 & 93 \\
    SingleAgent                & 52 & 48 & 42 & 65 & 70 & 70 & 38 \\
    AI Scientist-v2            & 65 & 22 & 28 & 28 & 42 & 48 & 32 \\
    GPT-5 ET                   & 14 & 24 & 29 & 22 & 38 & 42 & 17 \\
    PaperOrchestra*             & 75 & 81 & 74 & 81 & 76 & 83 & 79 \\
    \textbf{\method{}}   & \textbf{95} & \textbf{89} & \textbf{93} & \textbf{95} & \textbf{87} & \textbf{93} & \textbf{90} \\
    \bottomrule
  \end{tabular}%
  }
\end{table}

\paragraph{LLM and Human Peer Review. }
\label{sec:peer-review}
\begin{table}[h]
  \centering\footnotesize
  \caption{Simulated peer-review scores on five \method{}-generated papers vs.\ five 3D Gaussian Splatting papers accepted at CVPR or NeurIPS 2025. Each paper receives Soundness, Presentation, and Novelty on a 1--4 NeurIPS scale, plus an overall accept/reject score on the 1--6 scale (1 = strong reject, 6 = strong accept; 3.5 = borderline). Two judges score every paper: a multi-agent reviewer modeled on ScholarPeer~\citep{scholarpeer2026} and a graphics-area human reviewer blind to authorship. Means across all five papers per source.}
  \label{tab:peer-review}
  \resizebox{\linewidth}{!}{%
  \begin{tabular}{@{}llcccc@{}}
    \toprule
    Judge & Source & Soundness & Presentation & Novelty & \makecell{Overall\\(1--6)} \\
    \midrule
    \multirow{2}{*}{Multi-agent reviewer}
      & CVPR/NeurIPS 2025 (human GT) & 3.2 & 3.4 & 3.0 & 3.5 \\
      & \method{} (top 5)            & 3.4 & 3.6 & 3.4 & 3.6 \\
    \midrule
    \multirow{2}{*}{Human reviewer}
      & CVPR/NeurIPS 2025 (human GT)  & 3.6 & 3.8 & 3.1 & 4.1 \\
      & \method{} (top 5)             & 3.2 & 3.0 & 3.5 & 3.5 \\
    \bottomrule
  \end{tabular}%
  }
\end{table}

We review the top five \method{}-generated papers against five 3DGS papers accepted at CVPR or NeurIPS 2025 under the NeurIPS rubric (Soundness, Presentation, Novelty on 1--4; overall on 1--6, 3.5 = borderline). Each paper is scored independently by a multi-agent reviewer similar to ScholarPeer~\citep{scholarpeer2026}, with separate agents per review axis and a chair agent for aggregation, and by a graphics-area human reviewer blind to authorship. The multi-agent reviewer slightly favors \method{} on every axis (3.6 vs.\ 3.5 overall), consistent with the residual LLM-judge bias toward AI-generated content~\citep{zheng2023llmjudge}. The human reviewer reads the comparison differently: published papers score higher on Soundness and Presentation (3.6 and 3.8 vs.\ 3.2 and 3.0) but lower on Novelty (3.1 vs.\ 3.5), with the published set landing at clear accept (4.1) and \method{} at borderline (3.5). The split matches what the rest of the paper argues: verified plugins and the integrity audit close the soundness gap on numbers, but human authors still write better papers, while the machine-pace design-space sweep surfaces mechanisms human-paced research has not yet reached.

\begin{figure*}[!htbp]
  \centering
  \includegraphics[width=\linewidth]{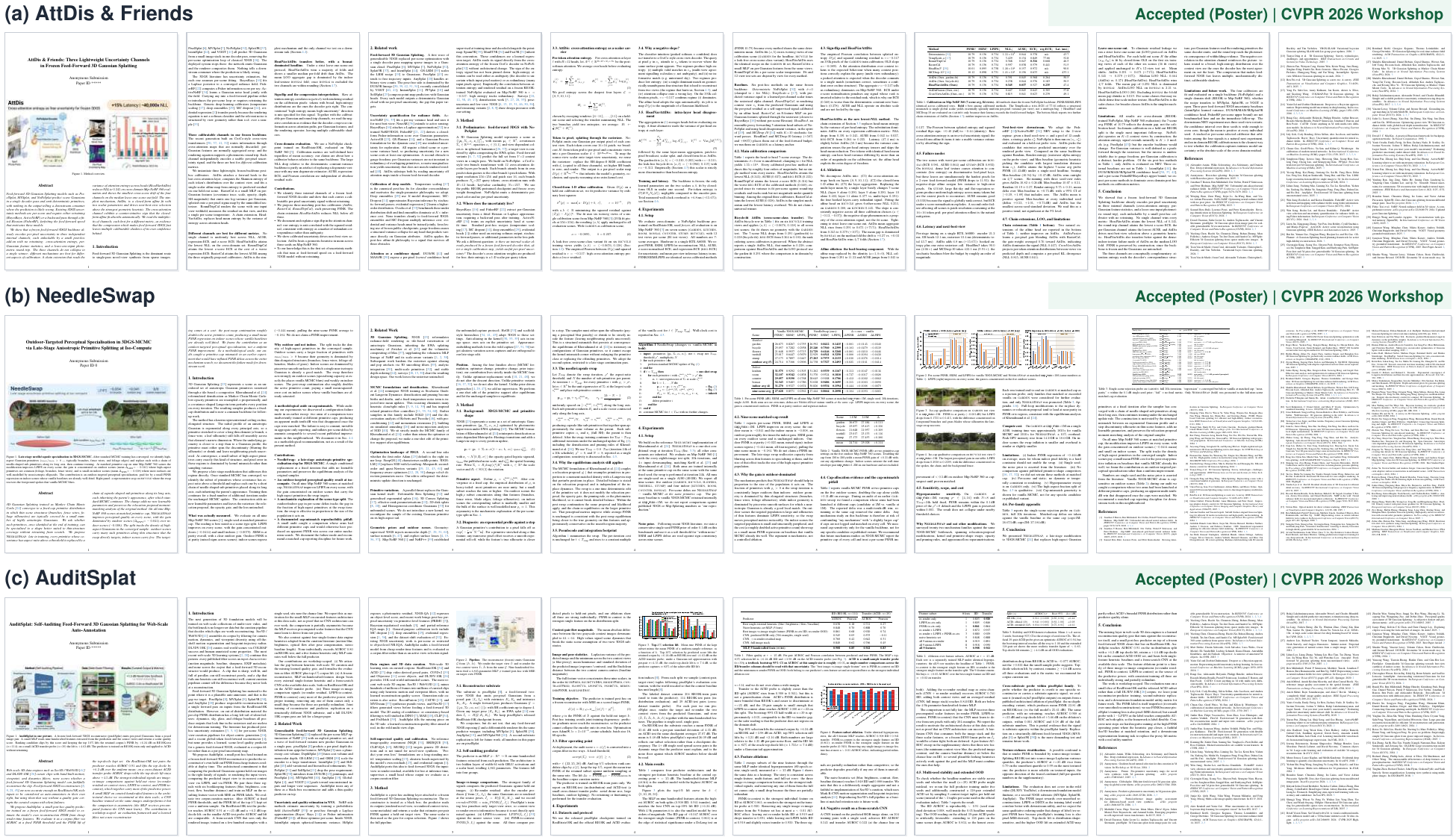}
  \caption{\textbf{Generated papers accepted at a CVPR 2026 workshop.} (a) \textbf{AttDis \& Friends}, (b) \textbf{NeedleSwap}, and (c) \textbf{AuditSplat}, all accepted. Each row shows pages 1--8 from left to right.}
  \label{fig:accepted-workshop-papers}
\end{figure*}

\paragraph{Real Workshop Acceptance and Ethical Protocol.}
\label{sec:workshop}
With prior approval from the workshop organizers and disclosure
of AI involvement, we submitted \method{}-generated
manuscripts to a CVPR 2026 workshop: \emph{AttDis \& Friends},
\emph{NeedleSwap}, and \emph{AuditSplat}. All three were accepted
as posters (Fig.~\ref{fig:accepted-workshop-papers}).
To address ethical concerns, we withdrew all submissions after
the peer-review experiment, using the acceptance decisions
solely to evaluate \method{} under real-world review.

\paragraph{Component Validation and Ablations. }
\label{sec:component-validation}

\begin{table*}[h]
\centering
\footnotesize
\setlength{\tabcolsep}{4pt}
\renewcommand{\arraystretch}{1.1}
\caption{Component ablations on the canonical pool. Each row removes one mechanism and reports the with-vs-without delta on a single metric.}
\label{tab:ablations}
\resizebox{\textwidth}{!}{%
\begin{tabular}{p{4.8cm}p{5.4cm}p{4.2cm}}
\toprule
\textbf{Component ablated} & \textbf{Without} & \textbf{With (full pipeline)} \\
\midrule
Stage 1: grounding (L/B-IDs) & 0/32 ideas cleared Stage 4 cascade & winner produced \\
Stage 2: Elo pre-selection & 24 GPU evals to winner & 6 GPU evals (20 rejected pre-GPU) \\
Stage 3: verified-plugin contract & 9/16 winners non-reproducible ($>1.0$\,dB) & 0/16 non-reproducible \\
Stage 4: cascade weights (uniform $w_s{=}1$) & smoke-test winner picked, 5th on multi-scene & multi-scene leader picked \\
Stage 4: per-$(B,L)$ Elo (global) & global champ $-1.6$\,dB on Nebula vs.\ specialist & per-target winners \\
Stage 4: verbal gradients (Reflection off) & 5 generations to stabilize, $2.4\times$ tokens & 2 generations \\
Stage 4: standing falsifier ($-30$ Elo off) & 5/16 winners fail held-out attacks & 0/16 fail \\
Stage 5: claim-backed audit gate & 16/16 manuscripts contain unbacked numbers & 0/16 unbacked \\
Stage 6: reviewer re-experimentation & avg simulated acceptance 2.2/6 & avg score 3.6/6 \\
\bottomrule
\end{tabular}}
\end{table*}

Each ablation removes one component, reuses the canonical-run winner and 
the same 32-idea / 16-paper pool for fair comparison, and reports the 
with-vs-without delta on a single concrete metric. Stage~5 carries the 
headline result: without the claim-backed verification gate, every 
ablated manuscript contained at least one number that did not trace to 
any log on disk, while the full pipeline produced 0 such cases. This 
closes the hallucination failure mode that AI Scientist drafts and 
successors leave open.

\paragraph{Portability and Taxonomy Ablation.}
Stages 1 and 3--6 are designed to operate across domains: agents build the literature corpus and codebase profile, select the evaluation metric, implement structured diffs, and run the tournament, audits, and reviewer loop against the target codebase and benchmark. Stage~2 supplies domain-specific ideation grounding; adapting its taxonomy requires expert effort, whose effect on search efficiency we examine below.
On the nebula target, taxonomy-guided and taxonomy-free
runs improved PSNR by $+3.78$ and $+3.45$ dB, respectively.
Removing the taxonomy increased token consumption from
47.5M to 85.5M (80\%). This single-target comparison
suggests that domain grounding improves search efficiency,
rather than being strictly necessary for execution.
Additional runs outside 3DGS with Stage~2 disabled are described in the Appendix.

\section{Conclusion}
\label{sec:conclusion}

\method{} shows that autonomous research in a live computer-vision
subfield is achievable when ideation operates atop verified
reproductions and every reported claim is bound to a logged training
run. From a one-line prompt or no input at all, \method{} produces
submission-ready 3DGS manuscripts that each beat their reproduced
target on PSNR. The research cycle (read, reproduce, extend, train,
ablate, write) that today takes a graduate student months compresses
to days of mostly autonomous compute, and a single research cluster
running \method{} sweeps design-space slices that hand-execution
would take years to cover. The bottleneck on autonomous research was
never model capability; it was the absence of four ingredients: a
domain framework, a verified baseline, iterative full-loop
refinement, and a binding from claim to measurement. Once those are
in place, the discovery rate is set by GPU throughput rather than
researcher time.

\paragraph{Broader impacts.}
\method{} lets small labs participate in live subfields without weeks of reimplementation. A tool that compresses months to days can flood peer review designed for human pace. The per-field cost of building a verified-reproduction backbone biases capability toward pre-engineered subfields and well-resourced labs. A bug in the integrity audit could mask hallucination instead of catching it. We address the third through Stage~5 audits and Mode~A archival; the first two need community norms. \method{} may reduce entry-level research opportunities, motivating AI research mentors and educational use of Mode~A logs to teach hypothesis formation, debugging, and critical evaluation of evidence. We release the system openly, recommend that derivative submissions be labelled and ship their experiment ledger, and discourage volume submissions. \method{} is a research instrument, not a paper-mill backend. Limitations and study-resource accounting are discussed in the Appendix.

\begin{ack}
This work was supported by institutional and advisor-provided research and computing resources at UC San Diego. The authors declare no competing interests.
\end{ack}
\bibliographystyle{splncs04}
\bibliography{references}

\begin{thebibliography}{10}
\providecommand{\url}[1]{\texttt{#1}}
\providecommand{\urlprefix}{URL }
\providecommand{\doi}[1]{https://doi.org/#1}

\bibitem{baek2025researchagent}
Baek, J., Jauhar, S.K., Cucerzan, S., Hwang, S.J.: {R}esearch{A}gent: Iterative
  research idea generation over scientific literature with large language
  models. In: Proceedings of the 2025 Conference of the Nations of the Americas
  Chapter of the Association for Computational Linguistics: Human Language
  Technologies (NAACL). pp. 6709--6738 (2025)

\bibitem{bagdasarian20253dgszip}
Bagdasarian, M.T., Knoll, P., Li, Y.H., Barthel, F., Hilsmann, A., Eisert, P.,
  Morgenstern, W.: {3DGS.zip}: A survey on {3D Gaussian Splatting} compression
  methods. Computer Graphics Forum  \textbf{44} (2025)

\bibitem{barron2022mipnerf360}
Barron, J.T., Mildenhall, B., Verbin, D., Srinivasan, P.P., Hedman, P.:
  Mip-{NeRF} 360: Unbounded anti-aliased neural radiance fields. In: IEEE/CVF
  Conference on Computer Vision and Pattern Recognition (CVPR). pp. 5470--5479
  (2022)

\bibitem{besta2024got}
Besta, M., Blach, N., Kubicek, A., Gerstenberger, R., Gianinazzi, L., Gajda,
  J., et~al.: Graph of thoughts: Solving elaborate problems with large language
  models. In: Proceedings of the AAAI Conference on Artificial Intelligence.
  vol.~38, pp. 17682--17690 (2024)

\bibitem{boiko2023coscientist}
Boiko, D.A., MacKnight, R., Kline, B., Gomes, G.: Autonomous chemical research
  with large language models. Nature  \textbf{624}(7992),  570--578 (2023).
  \doi{10.1038/s41586-023-06792-0}

\bibitem{buchanan1981dendral}
Buchanan, B.G., Feigenbaum, E.A.: Dendral and meta-dendral: Their applications
  dimension. In: Readings in artificial intelligence, pp. 313--322. Elsevier
  (1981)

\bibitem{clune2019ai}
Clune, J.: Ai-gas: Ai-generating algorithms, an alternate paradigm for
  producing general artificial intelligence. arXiv preprint arXiv:1905.10985
  (2019)

\bibitem{darcy2024marg}
D'Arcy, M., Hope, T., Birnbaum, L., Downey, D.: {MARG}: Multi-agent review
  generation for scientific papers. arXiv preprint arXiv:2401.04259  (2024)

\bibitem{feng2026internagent}
Feng, S., Ma, R., Yan, X., Fan, Y., Hu, Y., Huang, S., Zhang, S., Cao, Z.,
  Peng, T., Yuan, J., Guo, Z., Zhong, Z., Du, S., Wang, W., Shi, J., Zhou, Y.,
  He, X., Yu, Z., Yu, F., Zhan, B., Zheng, Q., Wu, J., Liu, M., Zhang, C., Hou,
  S., Li, S., Jiang, Y., Lou, W., Wang, L., Wang, Z., Wang, J., Xu, W., Deng,
  Y., Liu, D., Wang, Y., Zhang, W., Ling, F., Zhang, S., Wang, X., Zheng, S.,
  Huang, X., Sun, S., Hu, S., Ye, P., Song, C., Wang, B., He, C., Liu, Y., Li,
  X., Hou, Q., Chen, T., Yue, X., Wang, B., He, L., Lin, D., Zhou, B., Zhang,
  B., Bai, L.: Internagent-1.5: A unified agentic framework for long-horizon
  autonomous scientific discovery. arXiv preprint arXiv:2602.08990  (2026)

\bibitem{fernando2024promptbreeder}
Fernando, C., Banarse, D., Michalewski, H., Osindero, S., Rockt{\"a}schel, T.:
  Promptbreeder: Self-referential self-improvement via prompt evolution. In:
  Proceedings of the 41st International Conference on Machine Learning (ICML).
  Proceedings of Machine Learning Research, vol.~235, pp. 13481--13544 (2024)

\bibitem{gottweis2025towards}
Gottweis, J., Weng, W.H., Daryin, A., Tu, T., Palepu, A., Sirkovic, P.,
  Myaskovsky, A., Weissenberger, F., Rong, K., Tanno, R., et~al.: Towards an ai
  co-scientist. arXiv preprint arXiv:2502.18864  (2025)

\bibitem{scholarpeer2026}
Goyal, P., Parmar, M., Song, Y., Palangi, H., Pfister, T., Yoon, J.:
  Scholarpeer: A context-aware multi-agent framework for automated peer review.
  arXiv preprint arXiv:2601.22638  (2026)

\bibitem{hong2024metagpt}
Hong, S., Zhuge, M., Chen, J., Zheng, X., Cheng, Y., Wang, J., Zhang, C., Wang,
  Z., Yau, S.K.S., Lin, Z., Zhou, L., Ran, C., Xiao, L., Wu, C., Schmidhuber,
  J.: {M}eta{GPT}: Meta programming for a multi-agent collaborative framework.
  In: The Twelfth International Conference on Learning Representations (ICLR)
  (2024)

\bibitem{ifargan2025datatopaper}
Ifargan, T., Hafner, L., Kern, M., Alcalay, O., Kishony, R.: Autonomous
  {LLM}-driven research---from data to human-verifiable research papers. NEJM
  AI  \textbf{2}(1) (2025). \doi{10.1056/AIoa2400555}

\bibitem{jain2026splatifyreproducediscoverinnovate}
Jain, S., Gupta, K., Chandraker, M.: Splatify: Reproduce, discover, innovate!
  from papers and ideas to trainable 3dgs code (2026),
  \url{https://arxiv.org/abs/2610.09116}

\bibitem{jain2026nerfify}
Jain, S., Gupta, K., Gupta, K., Chandraker, M.: Nerfify: A multi-agent
  framework for turning nerf papers into code. arXiv preprint arXiv:2603.00805
  (2026)

\bibitem{jansen2025codescientist}
Jansen, P., Tafjord, O., Radensky, M., Siangliulue, P., Hope, T., Dalvi~Mishra,
  B., Majumder, B.P., Weld, D.S., Clark, P.: {C}ode{S}cientist: End-to-end
  semi-automated scientific discovery with code-based experimentation. In:
  Findings of the Association for Computational Linguistics: ACL 2025. pp.
  13370--13467 (2025)

\bibitem{jin2024agentreview}
Jin, Y., Zhao, Q., Wang, Y., Chen, H., Zhu, K., Xiao, Y., Wang, J.:
  {A}gent{R}eview: Exploring peer review dynamics with {LLM} agents. In:
  Proceedings of the 2024 Conference on Empirical Methods in Natural Language
  Processing (EMNLP). pp. 1208--1226 (2024)

\bibitem{kerbl20233dgs}
Kerbl, B., Kopanas, G., Leimk{\"u}hler, T., Drettakis, G.: {3D} {G}aussian
  splatting for real-time radiance field rendering. ACM Transactions on
  Graphics  \textbf{42}(4) (Jul 2023). \doi{10.1145/3592433}

\bibitem{kon2025curie}
Kon, P.T.J., Liu, J., Ding, Q., Qiu, Y., Yang, Z., Huang, Y., Srinivasa, J.,
  Lee, M., Chowdhury, M., Chen, A.: {Curie}: Toward rigorous and automated
  scientific experimentation with {AI} agents. arXiv preprint arXiv:2502.16069
  (2025)

\bibitem{lee2025mindevolution}
Lee, K.H., Fischer, I., Wu, Y.H., Marwood, D., Baluja, S., Schuurmans, D.,
  Chen, X.: Evolving deeper {LLM} thinking. arXiv preprint arXiv:2501.09891
  (2025)

\bibitem{li2024mlrcopilot}
Li, R., Patel, T., Wang, Q., Wang, Q., Du, X.: {MLR-Copilot}: Autonomous
  machine learning research based on large language model agents. In: NAACL
  (2025)

\bibitem{liu2024eoh}
Liu, F., Tong, X., Yuan, M., Lin, X., Luo, F., Wang, Z., Lu, Z., Zhang, Q.:
  Evolution of heuristics: Towards efficient automatic algorithm design using
  large language model. In: Proceedings of the 41st International Conference on
  Machine Learning (ICML). Proceedings of Machine Learning Research, vol.~235,
  pp. 32201--32223 (2024)

\bibitem{liu2024aigs}
Liu, Z., Liu, K., Zhu, Y., Lei, X., Yang, Z., Zhang, Z., Li, P., Liu, Y.:
  {AIGS}: Generating science from {AI}-powered automated falsification. arXiv
  preprint arXiv:2411.11910  (2024)

\bibitem{lu2024aiscientist}
Lu, C., Lu, C., Lange, R.T., Foerster, J., Clune, J., Ha, D.: The {AI}
  {S}cientist: Towards fully automated open-ended scientific discovery. arXiv
  preprint arXiv:2408.06292  (2024)

\bibitem{bran2024chemcrow}
M.~Bran, A., Cox, S., Schilter, O., Baldassari, C., White, A.D., Schwaller, P.:
  Augmenting large language models with chemistry tools. Nature Machine
  Intelligence  \textbf{6},  525--535 (2024). \doi{10.1038/s42256-024-00832-8}

\bibitem{mildenhall2019llff}
Mildenhall, B., Srinivasan, P.P., Ortiz-Cayon, R., Kalantari, N.K.,
  Ramamoorthi, R., Ng, R., Kar, A.: Local light field fusion: Practical view
  synthesis with prescriptive sampling guidelines. ACM Transactions on Graphics
  (TOG)  (2019)

\bibitem{nguyen2025dwtgs}
Nguyen, H., Li, R., Le, A., Nguyen, T.: Dwtgs: Rethinking frequency
  regularization for sparse-view 3d gaussian splatting (2025),
  \url{https://arxiv.org/abs/2507.15690}

\bibitem{zhang2025novelseek}
{NovelSeek Team}, Zhang, B., Feng, S., Yan, X., Yuan, J., Yu, Z., He, X.,
  Huang, S., Hou, S., Nie, Z., Wang, Z., Liu, J., Ma, R., Peng, T., Ye, P.,
  Zhou, D., Zhang, S., Wang, X., Zhang, Y., Li, M., Tu, Z., Yue, X., Ouyang,
  W., Zhou, B., Bai, L.: {NovelSeek}: When agent becomes the
  scientist---building closed-loop system from hypothesis to verification.
  arXiv preprint arXiv:2505.16938  (2025)

\bibitem{novikov2025alphaevolve}
Novikov, A., V{\~u}, N., Eisenberger, M., Dupont, E., Huang, P.S., Wagner,
  A.Z., Shirobokov, S., Kozlovskii, B., Ruiz, F.J., Mehrabian, A., et~al.:
  Alphaevolve: A coding agent for scientific and algorithmic discovery. arXiv
  preprint arXiv:2506.13131  (2025)

\bibitem{qian2024chatdev}
Qian, C., Liu, W., Liu, H., Chen, N., Dang, Y., Li, J., Yang, C., Chen, W., Su,
  Y., Cong, X., Xu, J., Li, D., Liu, Z., Sun, M.: {C}hat{D}ev: Communicative
  agents for software development. In: Proceedings of the 62nd Annual Meeting
  of the Association for Computational Linguistics (ACL), Volume 1: Long
  Papers. pp. 15174--15186 (2024)

\bibitem{ren2026fastgs}
Ren, S., Wen, T., Fang, Y., Lu, B.: {FastGS}: Training {3D Gaussian Splatting}
  in 100 seconds. In: IEEE/CVF Conference on Computer Vision and Pattern
  Recognition (CVPR) (2026), highlight

\bibitem{romera2024funsearch}
Romera-Paredes, B., Barekatain, M., Novikov, A., Balog, M., Kumar, M.P.,
  Dupont, E., Ruiz, F.J.R., Ellenberg, J.S., Wang, P., Fawzi, O., Kohli, P.,
  Fawzi, A.: Mathematical discoveries from program search with large language
  models. Nature  \textbf{625}(7995),  468--475 (2024).
  \doi{10.1038/s41586-023-06924-6}

\bibitem{schmidgall2025agentrxiv}
Schmidgall, S., Moor, M.: {A}gent{R}xiv: Towards collaborative autonomous
  research. arXiv preprint arXiv:2503.18102  (2025)

\bibitem{schmidgall2025agentlab}
Schmidgall, S., Su, Y., Wang, Z., Sun, X., Wu, J., Yu, X., Liu, J., Moor, M.,
  Liu, Z., Barsoum, E.: {Agent Laboratory}: Using {LLM} agents as research
  assistants. arXiv preprint arXiv:2501.04227  (2025)

\bibitem{si2024can}
Si, C., Yang, D., Hashimoto, T.: Can llms generate novel research ideas? a
  large-scale human study with 100+ nlp researchers. arXiv preprint
  arXiv:2409.04109  (2024)

\bibitem{song2026paperorchestra}
Song, Y., Song, Y., Pfister, T., Yoon, J.: {PaperOrchestra}: A multi-agent
  framework for automated {AI} research paper writing. arXiv preprint
  arXiv:2604.05018  (2026)

\bibitem{starace2025paperbench}
Starace, G., Jaffe, O., Sherburn, D., Aung, J., Chan, J.S., Maksin, L., Dias,
  R., Mays, E., Kinsella, B., Thompson, W., Heidecke, J., Glaese, A.,
  Patwardhan, T.: {P}aper{B}ench: Evaluating {AI}'s ability to replicate {AI}
  research. In: Proceedings of the 42nd International Conference on Machine
  Learning (ICML). Proceedings of Machine Learning Research, vol.~267, pp.
  56843--56873 (2025)

\bibitem{swanson2025virtuallab}
Swanson, K., Wu, W., Bulaong, N.L., Pak, J.E., Zou, J.: The virtual lab of {AI}
  agents designs new {SARS}-{C}o{V}-2 nanobodies. Nature  \textbf{646}(8085),
  716--723 (2025). \doi{10.1038/s41586-025-09442-9}

\bibitem{tancik2023nerfstudio}
Tancik, M., Weber, E., Ng, E., Li, R., Yi, B., Kerr, J., Wang, T.,
  Kristoffersen, A., Austin, J., Salahi, K., Ahuja, A., McAllister, D.,
  Kanazawa, A.: Nerfstudio: A modular framework for neural radiance field
  development. In: ACM SIGGRAPH Conference Proceedings (2023)

\bibitem{tang2025airesearcher}
Tang, J., Xia, L., Li, Z., Huang, C.: {AI}-researcher: Autonomous scientific
  innovation. arXiv preprint arXiv:2505.18705  (2025)

\bibitem{wang2024mineru}
Wang, B., Xu, C., Zhao, X., Ouyang, L., Wu, F., Zhao, Z., Xu, R., Liu, K., Qu,
  Y., Shang, F., Zhang, B., Wei, L., Sui, Z., Zhu, W., Liu, K., Wu, R., Jin,
  C., Peng, B., Dai, Y., He, C.: {MinerU}: An open-source solution for precise
  document content extraction. arXiv preprint arXiv:2409.18839  (2024)

\bibitem{wang2024openhands}
Wang, X., Li, B., Song, Y., Xu, F.F., Tang, X., Zhuge, M., et~al.: {OpenHands}:
  An open platform for {AI} software developers as generalist agents. arXiv
  preprint arXiv:2407.16741  (2024)

\bibitem{wei2025agenticscience}
Wei, J., Yang, Y., Zhang, X., Chen, Y., Zhuang, X., Gao, Z., Zhou, D., Wang,
  G., Gao, Z., Cao, J., Qiu, Z., He, X., Zhang, Q., You, C., Zheng, S., Ding,
  N., Ouyang, W., Dong, N., Cheng, Y., Sun, S., Bai, L., Zhou, B.: From {AI}
  for science to agentic science: A survey on autonomous scientific discovery.
  arXiv preprint arXiv:2508.14111  (2025)

\bibitem{weng2024cycleresearcher}
Weng, Y., Zhu, M., Bao, G., Zhang, H., Wang, J., Zhang, Y., Yang, L.:
  Cycleresearcher: Improving automated research via automated review. arXiv
  preprint arXiv:2411.00816  (2024)

\bibitem{wiki:automated_mathematician}
{Wikipedia contributors}: Automated {M}athematician --- {W}ikipedia{,} the free
  encyclopedia (2026),
  \url{https://en.wikipedia.org/wiki/Automated_Mathematician}, [Online;
  accessed 4-May-2026]

\bibitem{wu2024autogen}
Wu, Q., Bansal, G., Zhang, J., Wu, Y., Li, B., Zhu, E., Jiang, L., Zhang, X.,
  Zhang, S., Liu, J., Awadallah, A.H., White, R.W., Burger, D., Wang, C.:
  {A}uto{G}en: Enabling next-gen {LLM} applications via multi-agent
  conversation. In: Conference on Language Modeling (COLM) (2024)

\bibitem{yamada2025ai}
Yamada, Y., Lange, R.T., Lu, C., Hu, S., Lu, C., Foerster, J., Clune, J., Ha,
  D.: The ai scientist-v2: Workshop-level automated scientific discovery via
  agentic tree search. arXiv preprint arXiv:2504.08066  (2025)

\bibitem{yan2024msaa3dgs}
Yan, Z., Low, W.F., Chen, Y., Lee, G.H.: Multi-scale 3d gaussian splatting for
  anti-aliased rendering. In: Proceedings of the IEEE/CVF Conference on
  Computer Vision and Pattern Recognition (CVPR). pp. 20923--20931 (2024)

\bibitem{yang2024sweagent}
Yang, J., Jimenez, C.E., Wettig, A., Lieret, K., Yao, S., Narasimhan, K.,
  Press, O.: {SWE}-agent: Agent-computer interfaces enable automated software
  engineering. In: Advances in Neural Information Processing Systems (NeurIPS).
  vol.~37, pp. 50528--50652 (2024)

\bibitem{ye2024reevo}
Ye, H., Wang, J., Cao, Z., Berto, F., Hua, C., Kim, H., Park, J., Song, G.:
  {R}e{E}vo: Large language models as hyper-heuristics with reflective
  evolution. In: Advances in Neural Information Processing Systems 37 (NeurIPS)
  (2024)

\bibitem{ye2025gsplat}
Ye, V., Li, R., Kerr, J., Turkulainen, M., Yi, B., Pan, Z., Seiskari, O., Ye,
  J., Hu, J., Tancik, M., Kanazawa, A.: gsplat: An open-source library for
  {Gaussian Splatting}. Journal of Machine Learning Research  \textbf{26},
  1--15 (2025)

\bibitem{yuan2025dolphin}
Yuan, J., Yan, X., Zhang, B., Chen, T., Shi, B., Ouyang, W., Qiao, Y., Bai, L.,
  Zhou, B.: Dolphin: Moving towards closed-loop auto-research through thinking,
  practice, and feedback. In: Proceedings of the 63rd Annual Meeting of the
  Association for Computational Linguistics (ACL). pp. 21768--21789 (2025)

\bibitem{zheng2023llmjudge}
Zheng, L., Chiang, W.L., Sheng, Y., Zhuang, S., Wu, Z., Zhuang, Y., Lin, Z.,
  Li, Z., Li, D., Xing, E.P., Zhang, H., Gonzalez, J.E., Stoica, I.: Judging
  {LLM}-as-a-judge with {MT}-bench and chatbot arena. In: Advances in Neural
  Information Processing Systems 36 (NeurIPS) Datasets and Benchmarks Track
  (2023)

\bibitem{zhu2026paperbanana}
Zhu, D., Meng, R., Song, Y., Wei, X., Li, S., Pfister, T., Yoon, J.:
  Paperbanana: Automating academic illustration for ai scientists. arXiv
  preprint arXiv:2601.23265  (2026)

\end{thebibliography}







\end{document}